\documentclass{article}

\usepackage{PRIMEarxiv}

\usepackage[utf8]{inputenc} 
\usepackage[T1]{fontenc}    
\usepackage{hyperref}       
\usepackage{url}            
\usepackage{booktabs}       
\usepackage{amsfonts}       
\usepackage{nicefrac}       
\usepackage{microtype}      
\usepackage{lipsum}
\usepackage{fancyhdr}       
\usepackage{graphicx}       
\graphicspath{{media/}}     
\usepackage{mdframed}
\usepackage{arydshln}
\usepackage{placeins}
\usepackage{cite}

\title{RigoChat 2: an adapted language model to Spanish using a bounded dataset and reduced hardware}

\author{%
\begin{tabular}{ccc}
{\small Gonzalo Santamaría Gómez} & {\small  Guillem García Subies} & {\small Pablo Gutiérrez Ruiz} \\
{\small Mario González Valero} & {\small Natàlia Fuertes} & {\small Helena Montoro Zamorano} \\
{\small Carmen Muñoz Sanz} & {\small Leire Rosado Plaza} & {\small Nuria Aldama García} \\
{\small David Betancur Sánchez} & {\small Kateryna Sushkova} & {\small Marta Guerrero Nieto} \\
& {\small Álvaro Barbero Jiménez} & \\
& & \\
\end{tabular}
\\
\normalfont{Instituto de Ingeniería del Conocimiento} \\
\normalfont{\{name.firstsurname\}@iic.uam.es} \\
\normalfont{Madrid, Spain}
}

\begin{document}

\maketitle

\begin{abstract}
Large Language Models (LLMs) have become a key element of modern artificial intelligence, demonstrating the ability to address a wide range of language processing tasks at unprecedented levels of accuracy without the need of collecting problem-specific data. However, these versatile models face a significant challenge: both their training and inference processes require substantial computational resources, time, and memory. Consequently, optimizing this kind of models to minimize these requirements is crucial. In this article, we demonstrate that, with minimal resources and in a remarkably short time, it is possible to enhance a state-of-the-art model, specifically for a given language task, without compromising its overall capabilities using a relatively small pretrained LLM as a basis. Specifically, we present our use case, RigoChat 2, illustrating how LLMs can be adapted to achieve superior results in Spanish-language tasks.
\end{abstract}

\section{Introduction and Previous Work}

Training Large Language Models (LLMs) \cite{gpt-4, llama3} has become a cornerstone of Natural Language Processing (NLP), enabling sophisticated applications across various domains. These models are trained on vast corpora, often containing billions or even trillions of tokens, which allows them to capture complex linguistic patterns and generalize well across a wide range of tasks. However, their training process is computationally intensive and requires extensive hardware resources and time. This challenge is exacerbated by the need for powerful large data centers that support their training process, which is resource-demanding and time-consuming, making the creation of state-of-the-art models available to only a few organizations.

Efforts to optimize both LLMs training and inference have led to several promising approaches. One notable method involves the use of smaller, pre-trained language models (SLMs) as a foundation for fine-tuning on specific tasks \cite{allal2024SmolLM2}. This approach leverages the strengths of existing models while reducing the need for extensive training data and computational resources. A second approach includes the use of datasets as small as possible, maximizing statistical significance in those use cases to which the model will be exposed \cite{zhou2023limaalignment, penedo2024finewebdatasetsdecantingweb}.

Recent advancements have also seen the emergence of techniques aimed at reducing the computational requirements of LLMs during training. For example, LoRA \cite{hu2021loralowrankadaptationlarge} enables fine-tuning of LLMs using additive updates in the form of low-rank matrices, which significantly reduce the number of parameters that need to be updated, leading to a reduced memory usage. QLoRA \cite{dettmers2023qloraefficientfinetuningquantized} builds upon LoRA by further compressing the model through quantization of the original weights and keeping the low-rank matrices at full precision. 

Some libraries and research efforts such as llama.cpp \cite{llamacpp} include several methods to quantize and serve LLMs in a bounded hardware for inference. There are also more efficient computational algorithms to implement the self-attention operation, making use of specialized hardware features, such as FlashAttention \cite{dao2022flashattentionfastmemoryefficientexact, dao2023flashattention2fasterattentionbetter, shah2024flashattention3fastaccurateattention}, which can save computation efforts in both the training phase and inference time.

In summary, even though the training and inference of LLMs remains resource-intensive, various strategies have been developed to optimize these processes. These include the use of smaller datasets, model distillation such as Llama-3.1 \cite{llama3}, parameter-efficient methods like LoRA or QLoRA, or smaller models as a basis. Each of these approaches addresses different aspects of the computational burden, offering a range of solutions to make LLMs more accessible and efficient for a broader range of applications.

This work builds upon existing methodologies to demonstrate that, even with limited resources, state-of-the-art models can be aligned to specific use cases while maintaining general performance. The approach utilizes a 7-billion-parameter pre-trained LLM as a foundation. To illustrate this concept, the case of RigoChat 2 is presented, an Spanish-aligned model showcasing how existing LLMs can be adapted to improve their performance in Spanish-language tasks.

Before delving into subsequent sections, it is essential to emphasize that this work involved significant efforts in data collection and quality assessment. Extensive research has been conducted on data collection methodologies, as highlighted in studies such as \cite{grandury2024somos600mprojectgeneratingnlp}, where they also focused on obtaining data resources in Spanish. Our training process relied on the direct preference optimization algorithm (DPO) \cite{rafailov2024directpreferenceoptimizationlanguage}, which requires a dataset that reflects human judgment. However, this approach becomes prohibitively expensive when managing large volumes of data. Consequently, the development of automated methods to simulate human evaluation—particularly those leveraging models—emerges as a highly valuable alternative. Several studies \cite{kim2024prometheus2opensource, lee2024checkevalrobustevaluationframework, wei2024rethinkinggenerativelargelanguage, hu2024rankpromptstepbystepcomparisonsmake, huang2024selfevaluationlargelanguagemodel, wang2024usercentricmultiintentbenchmarkevaluating, risch2021semanticanswersimilarityevaluating, gritta2024humanrankevalautomaticevaluationlms} have addressed this challenge. Building on these contributions, we propose a methodology to generate automated evaluations based on LLMs, aiming to align them with assessments conducted by human experts (computational linguists).

Bearing this in mind, the contributions of this paper are the following:

\begin{itemize}
    \item We propose a \textbf{methodology for collecting and quality-filtering} open conversational corpora.
    \item A new technique to \textbf{augment conversation threads} in conversational datasets using LLMs is introduced.
    \item A review and experimental analysis of \textbf{automatic evaluation methods for LLM responses in the Spanish language} is carried out.
    \item As a result of the previous steps, a \textbf{high-quality conversational dataset for preferences learning in the Spanish language} is produced.
    \item By making use of this dataset we create the \textbf{RigoChat 2} model, \textbf{an alignment of an open-weights state-of-the-art LLM to the Spanish language}, producing better performance in this language.
\end{itemize}

\section{Training Data: Preparation and Methodology}

The core contribution of our work is a methodology for collecting and processing high-quality data to align a Large Language Model (LLM) to the Spanish language. This process involves a series of steps that progressively refine and augment a large corpus of Spanish conversations to produce a high-quality dataset suitable for preference learning (DPO). The steps are as follows:

\begin{enumerate}
    \item \textbf{Collecting data:} Aggregating a large-scale corpus of Spanish conversations from various public and proprietary sources.
    \item \textbf{Filtering high-quality Conversations (HQ):} Applying quality criteria to filter out low-quality and inappropriate conversations, resulting in a \textbf{High Quality dataset}.
    \item \textbf{Augmenting conversations (HQ+):} Using a batch of LLM to generate alternative responses for each conversation thread, expanding the dataset and improving its quality.
    \item \textbf{Evaluating responses:} Developing and employing automatic evaluation methods to assess the quality of the generated responses from private and open-source corpora designed by computational linguists.
    \textbf{\item Generating preferences (PD):} making use of the automated evaluations, we produce a \textbf{Preferences Dataset} aimed to produce an alignment toward responses either annotated by humans or generated by high-performing models. 
\end{enumerate}

In a nutshell, given that reinforcement learning over \textbf{preference data} has been shown to be an effective method to improve performance and align LLMs, in particular using the DPO algorithm \cite{rafailov2024directpreferenceoptimizationlanguage}, our focus was to transform data resources to this kind of data format. This process was specifically applied to Spanish, although the underlying concepts can be generalized to other languages and use cases.

\subsection{Data Resources}

The initial dataset is compiled from a combination of public and proprietary sources. After collecting a large amount of data, several filtering and quality classification steps were applied. First, a language filter was implemented: if the texts had been pre-classified by language, this classification was used; otherwise, the langdetect library \cite{nakatani2010langdetect} was employed with a confidence threshold of 75\%. Next, data quality was assessed following the criteria described in \ref{quality_filter}. Additionally, corpora were classified based on the variety and nature of responses, as detailed in \ref{sources_classification}. Finally, specific considerations regarding the IIC private corpora were addressed, including a comparison with other resources.

Table \ref{tab:alldatasets} summarizes the analyzed sources and the number of samples retained after applying all the quality filters.

\begin{table}[h!]
\centering
\renewcommand{\arraystretch}{1.25}
\begin{tabular}{l|c|c|c|c}
    \hline
    \textbf{Name} & \textbf{Samples} & \textbf{Filtered samples} & \textbf{Quality} & \textbf{Type} \\
    \hline
    Evol Instruct Spanish \cite{Chen_MultilingualSIFT_Multilingual_Supervised_2023} & 59022 & 0 & M & 4 \\
    Instruct-Aira Dataset \cite{nicholas22aira} & 40946 & 0 & M & 4 \\
    ShareGPT \cite{RyokoAI_ShareGPT52K} & 1643 & 0 & M & 4 \\
    LMSYS-Chat-1M \cite{zheng2023lmsyschat1m} & 23758 & 0 & M & 4 \\
    Chatbot Arena Conversations Dataset \cite{zheng2023judging} & 519 & 0 & M & 4 \\
    OpenAssistant \cite{köpf2023openassistantconversationsdemocratizing} & 15438 & 5917 & H & 1 \\
    OpenHermes 2.5 Spanish \cite{OpenHermes_25_Spanish} & 1000000 & 0 & M & 4 \\
    M2Lingual \cite{maheshwary2024m2lingual} & 2550 & 0 & M & 4 \\
    Multilingual BioASQ-6B \cite{garcíaferrero2024medical} & 2744 & 0 & M & 2 \\
    AsistenciaRefugiados \cite{somosnlp2024asistenciarefugiados} & 21431 & 0 & M & 2 \\
    Wikihow in Spanish \cite{quispe2024wikihowes} & 113160 & 0 & M & 4 \\
    LingComp\_QA \cite{LingComp_QA} & 1004 & 1004 & H & 1 \\
    somos-clean-alpaca-es \cite{alpacaes} & 44932 & 0 & M & 2 \\
    Databricks Dolly \cite{dolly} & 15015 & 4598 & H & 2 \\
    MentorES \cite{mentores} & 10175 & 10175 & H & 2 \\
    Aya Dataset \cite{singh2024aya} & 3854 & 3854 & H & 2 \\
    RAG Multilingual \cite{ragmt} & 15018 & 15018 & H & 4 \\
    IIC Private Corpora & 88822 & 2102 & H & 2 \\
    \hdashline
    Total & 1460031 & 26646 & - & - \\
    \hline
\end{tabular}
\vspace{2mm}
\caption{All the collected Spanish dataset resources, their corresponding number of samples, the filtered high-quality samples and their quality rating.}
\label{tab:alldatasets}
\end{table}
\FloatBarrier

\begin{figure}[h!]
    \centering
    \includegraphics[width=0.95\textwidth]{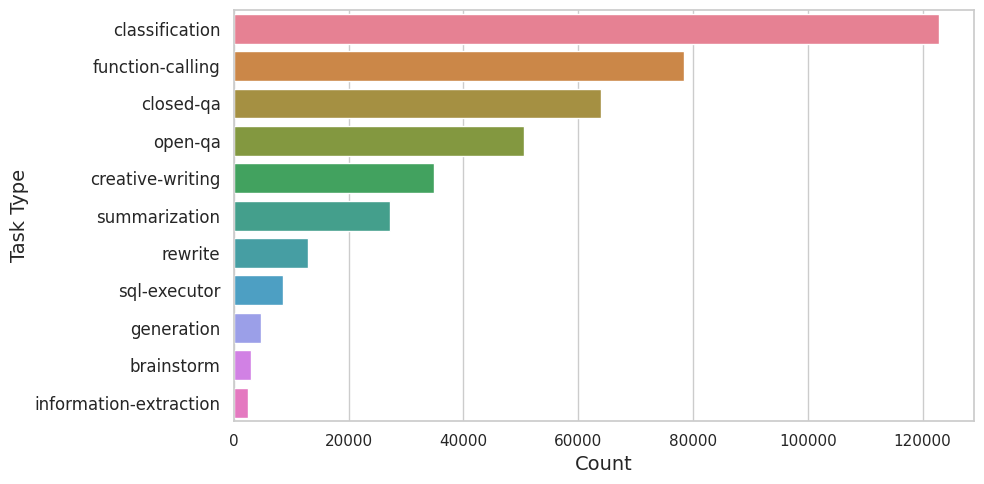}
    \caption{Task distribution estimation of the compilation of data resources.}
    \label{fig:alltaskd}
\end{figure}

Figure \ref{fig:alltaskd} illustrates the task distribution of all the collated the datasets as reported in their original sources.


\FloatBarrier



\subsection{Corpus-level quality filter (HQ)} \label{quality_filter}

To identify high-quality conversations, a corpus-level filtering process was conducted. Three computational linguists reviewed the datasets, assigning a quality rating based on the following criteria:

\subsubsection*{Low quality data}
\begin{itemize}
    \item \textbf{Grammatical and Linguistic Issues:} Texts with numerous grammatical errors, poorly executed translations, excessive use of anglicisms or fragmented sentences.
    \item \textbf{Inappropriate Language:} Texts containing insults, profanity, or other inappropriate language.
    \item \textbf{Encoding Errors:} Systematic issues such as incorrect characters or missing spaces.
    \item \textbf{Incomplete Information:} Instances where questions or answers are missing, responses include placeholders like \texttt{None}, or there is repetitive looping in responses.
\end{itemize}

\subsubsection*{Medium quality data}

\begin{itemize}
    \item Texts written in Spanish with only minor errors that do not significantly hinder comprehension. However, the accuracy of the content cannot be determined due to its technical or specific nature.
\end{itemize}

\subsubsection*{High-quality data}

\begin{itemize}
    \item Texts of exceptional quality, characterized by proper use of Spanish, relevant and engaging content, and reliable answers. This category includes texts with high conversational or contextual relevance, validated both by external reviewers and by assurances of correctness for highly technical responses provided by the creators of the datasets.
\end{itemize}

In some cases, such as OpenAssistant \cite{köpf2023openassistantconversationsdemocratizing}, human validations were already included. For these datasets, filtering was applied to retain only conversations with responses scoring above 70\%. Similarly, for the translated Dolly dataset \cite{dolly}, only the task types closed-qa and information-extraction were selected. After scoring each dataset, only those with a high quality rating were retained, filtering out datasets of medium or low quality. This resulted in a significantly reduced corpus, as shown in Table \ref{tab:alldatasets}.

\subsection{Data resources classification} \label{sources_classification}

Collected sources, along with the augmented data generated using the methodology outlined in the previous section, can be categorized based on whether they have been generated or evaluated by humans. This classification distinguishes four cases concerning the nature of the responses and their quality, as presented in Table \ref{tab:quality_criteria}.

\begin{table}[h!]
    \centering
    \renewcommand{\arraystretch}{1.25}
    \begin{tabular}{r|c|c}
        \hline
        \textbf{Data type} & \textbf{Expert annotation?} & \textbf{Variants validated?} \\
        \hline
        Type 1 & Yes & Yes  \\
        Type 2 & Yes & No  \\
        Type 3 & No & Yes  \\
        Type 4 & No & No  \\
        \hline
    \end{tabular}
    \vspace{2mm}
    \caption{Quality criteria classification for each thread in the dataset.}
    \label{tab:quality_criteria}
\end{table}
\FloatBarrier

The highest quality score (Type 1) corresponds to datasets manually created by human experts, with alternative responses either created or validated also by human experts. Second in quality (Type 2) drops the requirement of alternative responses to be manually checked, thus allowing for automated generation. Next (Type 3) are those datasets were all responses were created automatically, including the main response, but all of them were checked by a human. The lowest quality standard (Type 4) corresponds to fully automated and unrevised datasets.

According to this rating most of the collected datasets fall under Type 4. Only conversations of Types 1 and 3 are useful to apply DPO, as they provide a meaningful learning signal for the LLM. The Type 2 conversations will be automatically converted into Type 1 by augmenting the data, as described in the following sections, by assuming that the annotated answer is always more desirable than the generated alternatives.

\subsection{Data Augmentation (HQ+)} \label{data_augmented}

The high-quality dataset (HQ) was further augmented by generating alternative responses for each conversation thread using LLMs. Each conversation is segmented into multiple threads, corresponding to the number of turns it contains. For instance, a conversation with $n$ turns between a user and an assistant, represented as $c_n = \{u_1, a_1, \dots, u_n, a_n\}$, can be decomposed into $n$ separate \textbf{threads}:

\begin{itemize}
    \item $t_1$: $\{u_1, a_1\}$
    \item $t_2$: $\{u_1, a_1, u_2, a_2\}$
    \item $\cdots$
    \item $t_n$: $\{u_1, a_1, \dots, u_n, a_n\}$
\end{itemize}

Each thread $t_i$ can be further divided into input-output components:
\begin{itemize}
    \item Input: the chat history up to the user's latest query, $\{u_1, a_1, \dots, u_i\}$.
    \item Output: the assistant's most recent response, $a_i$.
\end{itemize}

For each thread $t_i$, we generated alternative responses using multiple LLMs. Each LLM $j$ was provided with the thread input, thereby generating an alternative response $a_i^j$. The resulting augmented thread is represented as:

\[ t_i = \{u_1, a_1, \dots, u_i, \{a_i, a_i^1, a_i^2, \dots, a_i^m\}\} \]

where $m$ is the number of deployed models, $a_i$ is the original response, and $a_i^j$ corresponds to the response generated by model $j$. The models used to generate new responses were the following: RigoChat-7b-v1 (private model), Qwen2-7B-Instruct \cite{qwen2}, 
Llama-3.1-8B-Instruct \cite{llama3}, gpt-3.5-turbo-0125 and gpt-4o-mini \cite{gpt-4o-mini}, stablelm-2-12b-chat \cite{bellagente2024stable} and gemma-2-9b-it \cite{gemmateam2024gemma2improvingopen}.

\subsection{Automating response scoring}

In order to create preference scores for the generated data without the need of human intervention we conducted an analysis of different automated metrics, with the objective to find a low-cost method that provided scores as close as possible to human judgment. Various types of automated evaluators were tested, including approaches based on metrics previously discussed in the state of the art and LLM-based methods as defined in the preceding section:

\begin{itemize}
    \item Open-source LLMs: Llama 3.1 \cite{llama3}.
    \item Commercial LLMs: GPT-4o \cite{gpt-4o}.
    \item Specialized evaluation models: Prometheus \cite{kim2024prometheus2opensource}.
    \item Metrics based on semantic similarity: SAS \cite{risch2021semanticanswersimilarityevaluating}.
    \item Ensembles combining several of these strategies. Specifically, an attempt was made to average several of these strategies to see if any improvement could be achieved.

\end{itemize}

Figure \ref{fig:aqa_prompts} shows the prompt that was used to transform an LLM into an evaluator with the same criteria as humans when addressing Abstractive Question-Answering tasks.

\begin{figure}[h!]
\begin{mdframed}
\textbf{LLM Evaluation System Prompt in Abstractive Question-Answering}: \\
\\
\\
Evaluarás la calidad de una respuesta generativa comparándola con una respuesta de referencia, basándote en un contexto dado y un histórico de la conversación. La evaluación será en una escala de 1 a 5, donde 1 es la peor calificación y 5 es la mejor. A continuación, se presentan las definiciones de cada puntuación: \\
\\
\\
- \textbf{Score 1:} La respuesta generada es irrelevante, incorrecta o no guarda ninguna relación con la pregunta, el contexto o la respuesta de referencia. La información es confusa o errónea.\\
\\
- \textbf{Score 2:} La respuesta generada es parcialmente relevante pero contiene errores significativos o malinterpretaciones en comparación con la respuesta de referencia. Muestra una comprensión limitada del contexto y no responde completamente a la pregunta.\\
\\
- \textbf{Score 3:} La respuesta generada es relevante y correcta en términos generales, pero carece de profundidad o detalle comparada con la respuesta de referencia. Cumple con los requisitos básicos, pero no aporta información adicional ni matices importantes.\\
\\
- \textbf{Score 4:} La respuesta generada es precisa y detallada, demostrando una buena comprensión del contexto y es comparable en calidad a la respuesta de referencia. Responde a la pregunta de manera completa y proporciona información adicional útil, aunque puede faltar algún pequeño detalle.\\
\\
- \textbf{Score 5:} La respuesta generada es excepcionalmente clara, completa y precisa. Demuestra una comprensión profunda del contexto y responde a la pregunta de manera exhaustiva, proporcionando detalles y matices que enriquecen la respuesta, igualando o superando la calidad de la respuesta de referencia.\\
\\
\\
Tu evaluación debe seguir el siguiente formato: \textbf{\{"score": <integer>\}}.

\end{mdframed}
\caption{System prompt used for the LLM evaluation in Abstractive Question-Answering.}
\label{fig:aqa_prompts}
\end{figure}
\FloatBarrier

Table \ref{tab:accuracyeval} summarizes the accuracy of various automatic evaluators compared to human evaluations. The two-model ensemble combined the Llama-3.1 and SAS scores, while the three-model ensemble combined, in addition to those already mentioned, the Prometheus metric. The Quantized Llama-3.1-70B-Instruct model achieved the highest average accuracy of 0.452.

\begin{table}[h!]
\centering
\renewcommand{\arraystretch}{1.25}
\begin{tabular}{l|c|c|c|c|c|c}
\hline
\textbf{Model} & \textbf{Average} & \textbf{Score 1} & \textbf{Score 2} & \textbf{Score 3} & \textbf{Score 4} & \textbf{Score 5} \\
\hline
Prometheus & 0.276 & 0.8 & 0.25 & 0.29 & 0 & 0.04 \\
Normalized SAS & 0.33 & 0.7 & 0 & 0 & 0 & 0.95 \\
Llama-3.1-8B-Instruct & 0.404 & 0.3 & 0.25 & 0.25 & 0.58 & 0.64 \\
Ensemble of 3 models& 0.39 & 0.6 & 0.12 & 0.29 & 0.84 & 0.1 \\
Ensemble of 2 models& 0.364 & 0.27 & 0.22 & 0.07 & 0.65 & 0.61 \\
Quantized Llama-3.1-70B-Instruct & \textbf{0.452} & 0.5 & 0.71 & 0.09 & 0.53 & 0.43 \\
GPT-4o & 0.356 & 0.5 & 0.29 & 0.27 & 0.53 & 0.19 \\
\hline
\end{tabular}
\vspace{2mm}
\caption{Accuracy for each metric in each score definition.}
\label{tab:accuracyeval}
\end{table}
\FloatBarrier

Table \ref{tab:matches} shows the number of matches and correlation coefficients between the automatic evaluations and human evaluations. The Llama-3.1-8B-Instruct model had the highest number of matches (43 out of 90) and strong correlations.

\begin{table}[h!]
\centering
\renewcommand{\arraystretch}{1.25}
\begin{tabular}{l|c|c|c}
\hline
\textbf{Model} & \textbf{Matches} & \textbf{Pearson correlation} \cite{pearson} & \textbf{Kendall correlation} \cite{kendall1938measure} \\
\hline
Prometheus & 13 & 0.42 & 0.33 \\
Normalized SAS & 27 & 0.68 & 0.57 \\
Llama-3.1-8B-Instruct & 43 & 0.62 & 0.48 \\
Ensemble 3 Models & 27 & 0.62 & 0.45 \\
Ensemble 2 Models & 41 & 0.65 & 0.49 \\
Quantized Llama-3.1-70B-Instruct & 36 & 0.74 & 0.56 \\
GPT-4o & 31 & 0.73 & 0.59 \\
\hline
\end{tabular}
\vspace{2mm}
\caption{Matches and correlations between models and human evaluation.}
\label{tab:matches}
\end{table}
\FloatBarrier

\begin{figure}[h!]
\centering
\includegraphics[width=\textwidth]{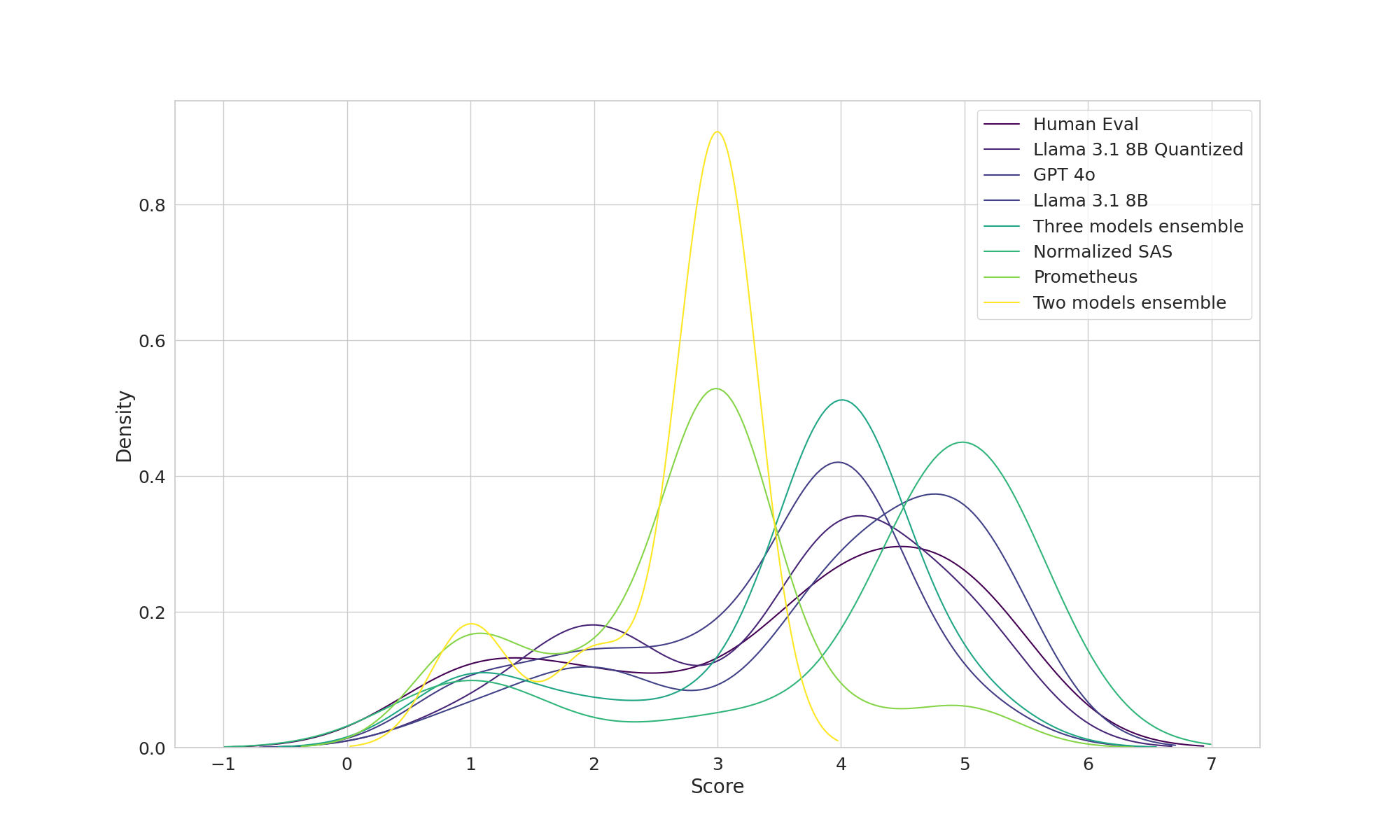}
\caption{Density comparisons across different metrics. Legend and color map are sorted based on distance to human evaluations.}
\label{fig:densityscores}
\end{figure}

Figure \ref{fig:densityscores} compares the score density distributions across the 90 example conversations, showing that the LLM-based evaluators, particularly Llama-3.1-based and GPT-4o, had the most similar distributions to human evaluations. The distributions are discrete, since all the evaluations have been characterized as 5 different scores, but the graph shows a KDE Plot with all the approximate distributions, ordered by their proximity to the human evaluation  (Human Eval), based on the Wasserstein distance \cite{Villani2009}. The Llama-3.1-8B-Instruct model was selected as the automated evaluator due to its strong performance in matching human evaluations and its efficiency.

\FloatBarrier

\subsection{IIC Private Corpora}

The data were created from plain text extracted from the web, encompassing information from various domains. To collect these data, texts were carefully selected from the chosen domains, and contexts were subsequently extracted to formulate corresponding questions and answers. Particular emphasis was placed on ensuring that the contexts were comprehensive and detailed.

All high-quality private corpora were generated and reviewed manually by humans following a specific annotation guide. This guide required computational linguists to generate question-answer pairs based on a given context. Each response was written by an annotator and subsequently reviewed by other annotators to ensure accuracy and consistency.

In addition to using high-quality sources, experiments were conducted to generate quality data for function calling, autonomous agents, and SQL execution tasks. These experiments aimed to address underrepresented use cases in Spanish. However, the data was ultimately discarded due to the lack of human review.

\subsection{Preferences Dataset (PD)}

Based on the filter and data augmentation criteria, we created a Preferences Dataset (PD). Only the Type 1, 2, and 3 conversations described in Table \ref{tab:quality_criteria} were retained. For the Type 2 conversations, which lack variants but were annotated by experts, we decided to use only the private sources of the IIC, although many others sources passed the quality filters. This decision was made to better represent the kind of task we wanted to put focus on: abstractive question answerin.

New responses were generated using the LLMs mentioned in \ref{data_augmented} under the assumption that the expert-annotated response is always better than any model-generated response, and the Llama-3.1-8B-Instruct model was used as evaluator to score these responses and sort them. Finally, the best and worst generated responses were selected, resulting in three pairs of preferences:

\begin{itemize}
    \item Expert-annotated is better than the best-scored generated response.
    \item Expert-annotated is better than the worst-scored generated response.
    \item The best-scored generated response is better than the worst-scored generated response.
\end{itemize}

Notably, the only case where humans are not involved is the latter, as both the evaluation and the responses are generated by models. In this way, the model is learning both from human feedback and from the outputs of a series of LLMs, arranged by their quality, resulting in a form of model distillation.

The final preference corpus comprises 21,975 conversations, including variants, suitable for both reinforcement learning algorithms and supervised fine-tuning. Table \ref{tab:preferences_datasets} details the sources and number of samples in the preference dataset.

\begin{table}[h!]
\centering
\renewcommand{\arraystretch}{1.25}
\begin{tabular}{l|c}
\hline
\textbf{Name} & \textbf{nº samples} \\
\hline
OpenAssistant \cite{köpf2023openassistantconversationsdemocratizing} & 16870 \\
IIC Private Corpora & 5105 \\
\hline
\end{tabular}
\vspace{2mm}
\caption{Preferences dataset after data augmentation.}
\label{tab:preferences_datasets}
\end{table}
\FloatBarrier

Table \ref{tab:split} provides the number of samples in the train and validation subsets of the preference dataset.

\begin{table}[h!]
\centering
\renewcommand{\arraystretch}{1.25}
\begin{tabular}{c|c}
\hline
\textbf{Split} & \textbf{nº samples} \\
\hline
Train & 21755 \\
Validation & 220 \\
\hline
\end{tabular}
\vspace{2mm}
\caption{Preferences dataset splitting.}
\label{tab:split}
\end{table}
\FloatBarrier

In summary, the process of deriving the preference dataset involved collecting a large-scale corpus, filtering it to retain high-quality conversations, augmenting it with alternative responses, and evaluating the quality of these responses. This iterative approach ensures that the final dataset is well-suited for preference learning and the alignment of LLMs to the Spanish language.

\section{Model}

Pursuing our objective of creating a model with limited computational resources, open-source models were utilized and aligned with the Preferences Dataset (PD) described in the previous section, using the Direct Preference Optimization (DPO) algorithm \cite{rafailov2024directpreferenceoptimizationlanguage}. Additionally, the final model was quantized to different precisions using llama.cpp \cite{llamacpp} to allow for efficient inference.

\subsection{Architecture}

RigoChat-7b-v2 is a model based on Qwen-2.5 \cite{qwen2, qwen2.5}. Specifically, it builds upon the Qwen2.5-7B-Instruct model and has been fine-tuned using DPO to enhance its performance in the Spanish language. We do not perform any modifications in the model architecture or its tokenizer.

\subsection{Training process}

The training details are provided in Table \ref{tab:dpotraining}. For this work, the Transformer Reinforcement Learning (TRL) library \cite{vonwerra2022trl} was employed. Specifically, the example script published for applying DPO was adapted to the dataset presented in Table \ref{tab:split}.

\begin{table}[h!]
\centering
\renewcommand{\arraystretch}{1.25}
\begin{tabular}{c|c}
\textbf{DPOConfig argument} & \textbf{Value} \\
\hline
num\_train\_epochs & $2$ \\
eval\_steps & $500$ \\
save\_steps & $100$ \\
save\_total\_limit & $5$ \\
per\_device\_train\_batch\_size & $1$ \\
per\_device\_eval\_batch\_size & $1$ \\
gradient\_accumulation\_steps & $16$ \\
learning\_rate & $5\cdot 10^{-6}$ \\
max\_length & $8192$ \\
max\_prompt\_length & $6656$ \\
gradient\_checkpointing & True \\
weight\_decay & $0.001$ \\
optim & rmsprop \\
lr\_scheduler\_type & cosine \\
\hline
\end{tabular}
\vspace{2mm}
\caption{Configuration for the DPO training. Parameters not shown here used  the default values of the library.}
\label{tab:dpotraining}
\end{table}
\FloatBarrier

To manage memory usage during training, the Parameter-Efficient Fine-Tuning (PEFT) framework \cite{peft} was employed, particularly the LoRA method \cite{hu2021loralowrankadaptationlarge}. The model configuration details are also available in Table \ref{tab:loraconfig}.

\begin{table}[h!]
\centering
\renewcommand{\arraystretch}{1.25}
\begin{tabular}{c|c}
\textbf{LoRA Config Parameter} & \textbf{Value} \\
\hline
r & $64$ \\
lora\_alpha & $16$ \\
lora\_dropout & $0.1$ \\
task\_type & CAUSAL\_LM \\
target\_modules & all-linear \\
use\_rslora \cite{kalajdzievski2023rankstabilizationscalingfactor} & True \\
\hline
\end{tabular}
\vspace{2mm}
\caption{Configuration for the PEFT model and LoRA hyperparameters. Parameters not shown here used  the default values of the library.}
\label{tab:loraconfig}
\end{table}
\FloatBarrier

\begin{figure}[h!]
    \centering
    \includegraphics[width=0.8\textwidth]{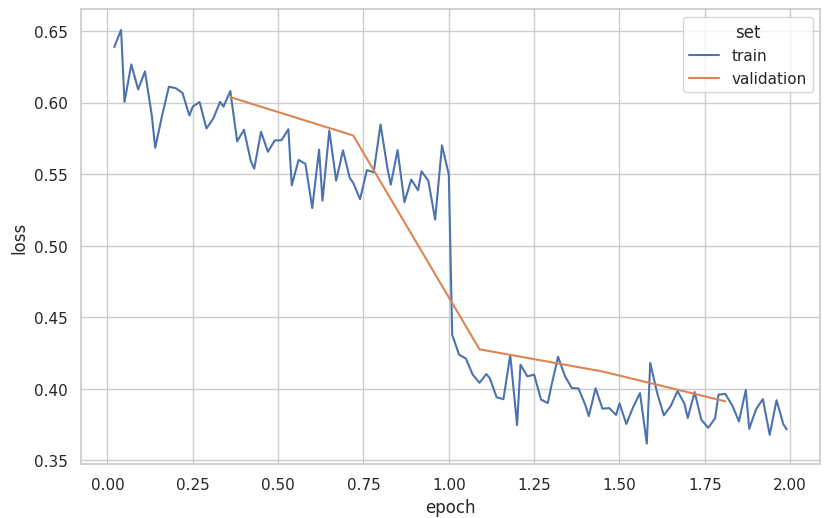}
    \caption{DPO Loss \cite{rafailov2024directpreferenceoptimizationlanguage} of the training process.}
    \label{fig:trainingprocess}
\end{figure}

The model was trained on a single NVIDIA A100 GPU with Tensor Core for $8.5$ hours, a minimal fraction of the compute power required when training a full LLM from scratch. However, and as discussed in subsequent sections, these results demonstrate that it is possible to align LLMs to a particular language, with limited hardware resources, while maintaining or slightly improving performance compared to state-of-the-art models across various use cases. Figure \ref{fig:trainingprocess} presents the evolution of the DPO loss during the training process. A noticeable reduction in error was observed during the second epoch; while this might hint some kind of overfitting to our preference data, further results will show that the performance of the model does in fact increase for other datasets.

\FloatBarrier

The algorithm demonstrated significant sensitivity to the learning rate, with catastrophic forgetting occurring at values above $10^{-5}$, rendering the experiment ineffective, while smaller values showed no substantial improvements. Additionally, experiments were conducted using different training approaches, including fine-tuning the base model Qwen2.5-7B with the high-quality instruction dataset from Table \ref{tab:alldatasets} before applying DPO, but results were inferior to starting directly from Qwen2.5-7B-Instruct. Other tests explored the impact of supervised learning alone, without DPO, as well as alternative state-of-the-art models. This section presents only the most successful approach, while a summary of additional experiments is provided in the evaluation section as ablation studies (refere to \ref{comparisons}).

\subsection{Quantizations}

Quantized versions of the trained model were also generated using the llama.cpp library \cite{llamacpp}, to allow for efficient inference. Most quantization methods rely on truncation, scaling, or similar strategies to convert bfloat16 precision weights into lower-precision weights. However, some techniques employ a small dataset to calibrate the final model. For those strategies we employed the best threads (responses validated by humans) from the Preferences Dataset described in \ref{tab:preferences_datasets}.

All quantization methods, along with their bit precision, types and size are detailed in Table \ref{tab:quantresults}. Quantizations requiring a calibration dataset are prefixed with an `i', indicating that an importance matrix was  be computed during the quantization process. This approach is particularly useful when the precision of the quantized model is too low. For more information on llama.cpp quantization methods, see \cite{quantypes}; for details on the importance matrix computation, refer to \cite{llamacppimatrix}.

\subsection{Model availability}

RigoChat 2 (\url{https://huggingface.co/IIC/RigoChat-7b-v2}) and all its quantized variants (\url{https://huggingface.co/IIC/RigoChat-7b-v2-GGUF}) are publicly available at Hugging Face under a permissive license suitable for research and non-commercial applications.

\section{Evaluations}

To assess the performance of RigoChat 2 and other competing LLMs, several high-quality corpora tailored to specific evaluation needs were developed:

\begin{itemize}
    \item \textbf{AQuAS \cite{aquas}:} A manually curated corpus created by two computational linguists to evaluate language models in the task of Abstractive Question-Answering in Spanish. It includes examples from domains such as finance, insurance, healthcare, law, and music.

    \item \textbf{RagQuAS \cite{ragquas}:} Another manually curated corpus developed by the same linguists to evaluate full RAG systems and language models in Abstractive Question-Answering tasks in Spanish. This corpus spans a wide range of domains, including hobbies, linguistics, pets, health, astronomy, customer service, cars, daily life, documentation, energy, skiing, fraud, gastronomy, languages, games, nail care, music, skating, first aid, recipes, recycling, complaints, insurance, tennis, transportation, tourism, veterinary, travel, and yoga.

    \item \textbf{CAM (private corpus):} This corpus consists of frequently asked questions (FAQs) sourced from consumer-related topics on the public websites of the Comunidad de Madrid (Autonomous Community of Madrid). The questions are categorized into three levels of degradation—E1, E2, and E3—intended to measure the LLMs’ ability to understand and effectively respond to poorly formulated queries caused by spelling errors, varying levels of colloquialism, and similar issues. This task also falls under the Abstractive Question-Answering category.

    \item \textbf{Shops (private corpus):} A multi-turn conversational corpus centered on policies from various clothing companies. The task involves Multi-turn Abstractive Question-Answering.

    \item \textbf{Insurance (private corpus):} Another multi-turn conversational corpus, this one focuses on policies from various insurance companies. It also involves Multi-turn Abstractive Question-Answering.
\end{itemize}

Each corpus includes the following columns: question, answer, and context(s) containing relevant information from which the model can derive the answer. In multi-turn tasks, a chat history is also provided. It is important to emphasize that the datasets used for model evaluation are distinct from those used for training in the IIC Private Corpora explained in Tables \ref{tab:alldatasets} and \ref{tab:preferences_datasets}. Among our private training corpora, they share only a similar distribution—not specific examples—with the \textbf{AQuAS}, \textbf{RagQuAS}, and \textbf{Shops} datasets.

To score all responses the Llama-3.1-8b-Instruct model evaluator with the approach described in Figure \ref{fig:aqa_prompts} was used. All corpora are private except for AQuAS and RagQuAS, which are publicly available and can serve as examples of the structure and content of the others.

\subsection{Comparison against other LLMs}

Table \ref{tab:evalresults} shows the results of evaluating RigoChat 2 and other relevant LLMs over the presented evaluation datasets. The Llama-3.1-8B-Instruct model was used as the evaluator and it is important to highlight that, due to the variability in LLMs' responses in real-world use cases—arising from factors such as generation parameters and system prompt configurations—we have estimated a margin of error of 0.5\% in these scores. This estimate was obtained by conducting repeated evaluations across various models.

\begin{table}[h!]
    \centering
    \renewcommand{\arraystretch}{1.2}
    \resizebox{\textwidth}{!}{%
        \begin{tabular}{r|c|c|c|c|c|c|c|c|c}
            \textbf{Model} & \textbf{Average} & \textbf{AQuAS} & \textbf{RagQuAS} & \textbf{CAM} & \textbf{CAM\_E1} & \textbf{CAM\_E2} & \textbf{CAM\_E3} & \textbf{Shops} & \textbf{Insurance}\\
            \hline
            RigoChat-7b-v2 & \textbf{79.55} & 82.52 & 79.10 & \textbf{78.91} & \textbf{79.17} & 76.73 & \textbf{78.23} & \textbf{80.79} & \textbf{81.04} \\
            GPT-4o \cite{gpt-4o, openai2024gpt4ocard} & 78.26 & \textbf{85.23} & 77.91 & 78.00 & 74.91 & 73.45 & 77.09 & 78.60 & 80.89 \\
            stablelm-2-12b-chat \cite{bellagente2024stable} & 77.74 & 78.88 & 78.21 & 77.82 & 78.73 & \textbf{77.27} & 74.73 & 77.03 & 79.26 \\
            Mistral-Small-Instruct \cite{mistral_small} & 77.29 & 80.56 & 78.81 & 77.82 & 75.82 & 73.27 & 73.45 & 78.25 & 80.36 \\
            Qwen2.5-7B-Instruct \cite{qwen2, qwen2.5} & 77.17 & 80.93 & 77.41 & 77.82 & 75.09 & 75.45 & 72.91 & 78.08 & 79.67 \\
            Llama-3.1-8B-Instruct \cite{llama3} & 76.55 & 81.87 & 80.50 & 72.91 & 73.45 & 75.45 & 71.64 & 77.73 & 78.88 \\
            GPT-4o-mini \cite{gpt-4o-mini} & 76.48 & 82.80 & 75.82 & 76.36 & 74.36 & 72.36 & 71.82 & 78.25 & 80.08 \\
            Phi-3.5-mini-instruct \cite{abdin2024phi3technicalreporthighly} & 76.38 & 81.68 & \textbf{81.09} & 75.82 & 74.73 & 71.45 & 70.36 & 77.43 & 78.45 \\
            gemma-2-9b-it \cite{gemmateam2024gemma2improvingopen} & 75.80 & 82.80 & 78.11 & 72.91 & 73.45 & 71.09 & 71.27 & 77.08 & 79.72 \\
            Ministral-8B-Instruct \cite{ministral} & 75.19 & 79.63 & 77.31 & 76.00 & 73.45 & 72.36 & 70.18 & 76.44 & 76.14 \\
            GPT-3.5-turbo & 74.78 & 80.93 & 73.53 & 76.73 & 72.55 & 72.18 & 69.09 & 75.63 & 77.64 \\
            Llama-2-7b-chat \cite{touvron2023llama2openfoundation} & 71.18 & 67.10 & 77.31 & 71.45 & 70.36 & 70.73 & 68.55 & 72.07 & 71.90 \\
            granite-3.0-8b-instruct \cite{granite} & 71.08 & 73.08 & 72.44 & 72.36 & 71.82 & 69.09 & 66.18 & 69.97 & 73.73 \\
            RigoChat-7b-v1 & 62.13 & 72.34 & 67.46 & 61.27 & 59.45 & 57.45 & 57.64 & 62.10 & 59.34 \\
            salamandra-7b-instruct \cite{salamandra} & 61.96 & 63.74 & 60.70 & 64.91 & 63.27 & 62.36 & 60.55 & 59.94 & 60.23 \\
            \hline
        \end{tabular}
    }
    \vspace{2mm}
    \caption{Evaluation results across different models and tasks in Spanish.}
    \label{tab:evalresults}
\end{table}
\FloatBarrier

\begin{figure}[h!]
    \centering
    \includegraphics[width=\textwidth]{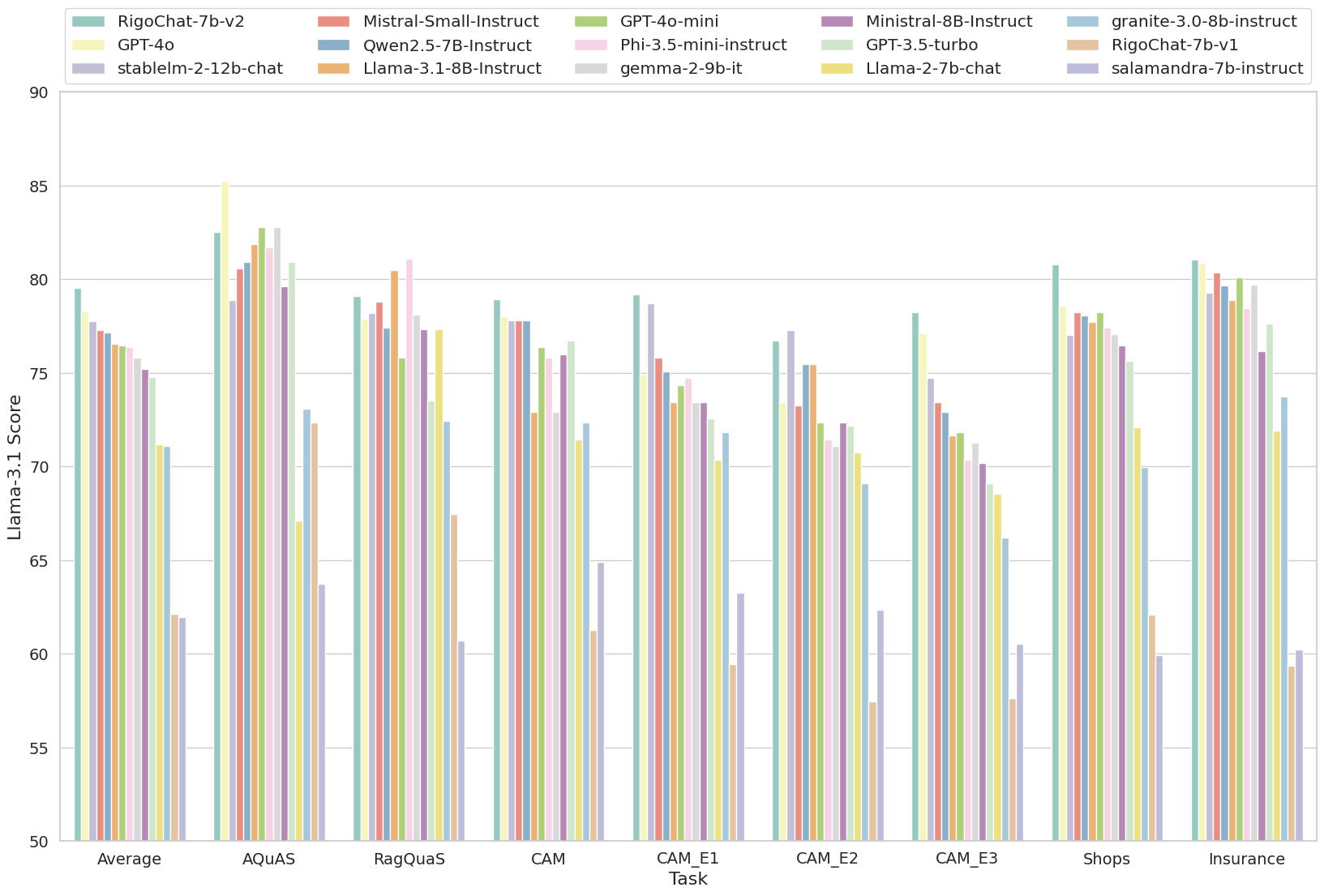}
    \caption{Bar Plot of the evaluation results. The figure is scaled from 50 to 90 in order to better appreciate the differences between all tested models.}
    \label{fig:evalresults}
\end{figure}

The information in Table \ref{tab:evalresults} is condensed in Figure \ref{fig:evalresults}. RigoChat-7b-v2 demonstrates a significant improvement over other state-of-the-art models of similar size, including the base model we used for training (Qwen2.5-7B-Instruct), and even surpassing more powerful models such as GPT-4o in these evaluations. This result highlights the feasibility of adapting LLMs to specific use cases with limited resources and reduced training time. Additionally, due to the complexity of the training dataset, these evaluations present certain challenges. Further improvements could likely be achieved if the training process was more closely aligned with the evaluated use cases.

\FloatBarrier

\begin{figure}[h!]
    \centering
    \includegraphics[width=\textwidth]{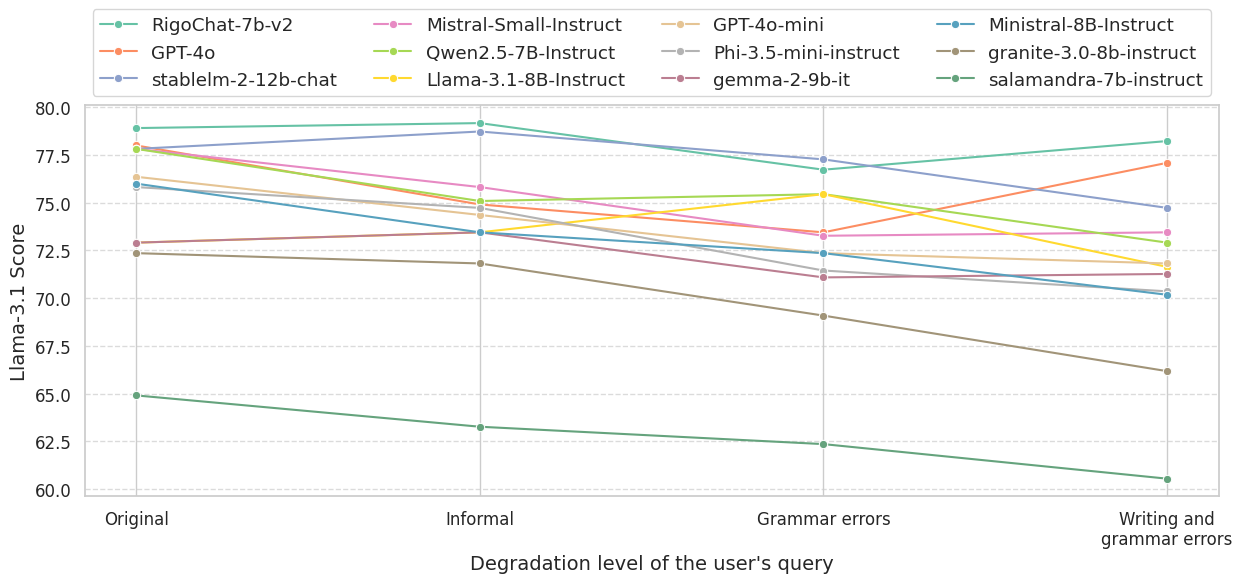}
    \caption{Evolution of LLM performance as a function of degradation.}
    \label{fig:degradation}
\end{figure}

A final comparison involves evaluating the sensitivity of language models to colloquialisms, jargon, or any type of noise introduced into user queries. For this purpose, the private CAM evaluation corpus is used, which quantifies different query variants with varying levels of degradation. Figure \ref{fig:degradation} illustrates the performance evolution of the models as noise increases in user queries. The general trend shows a decline in score as colloquialisms, jargon, and grammatical errors increase. However, RigoChat-7b-v2 demonstrates greater resilience, maintaining its performance more effectively than the other models.

\FloatBarrier

\subsection{Performance of RigoChat 2 quantizations }

\begin{figure}[h!]
    \centering
    \includegraphics[width=\textwidth]{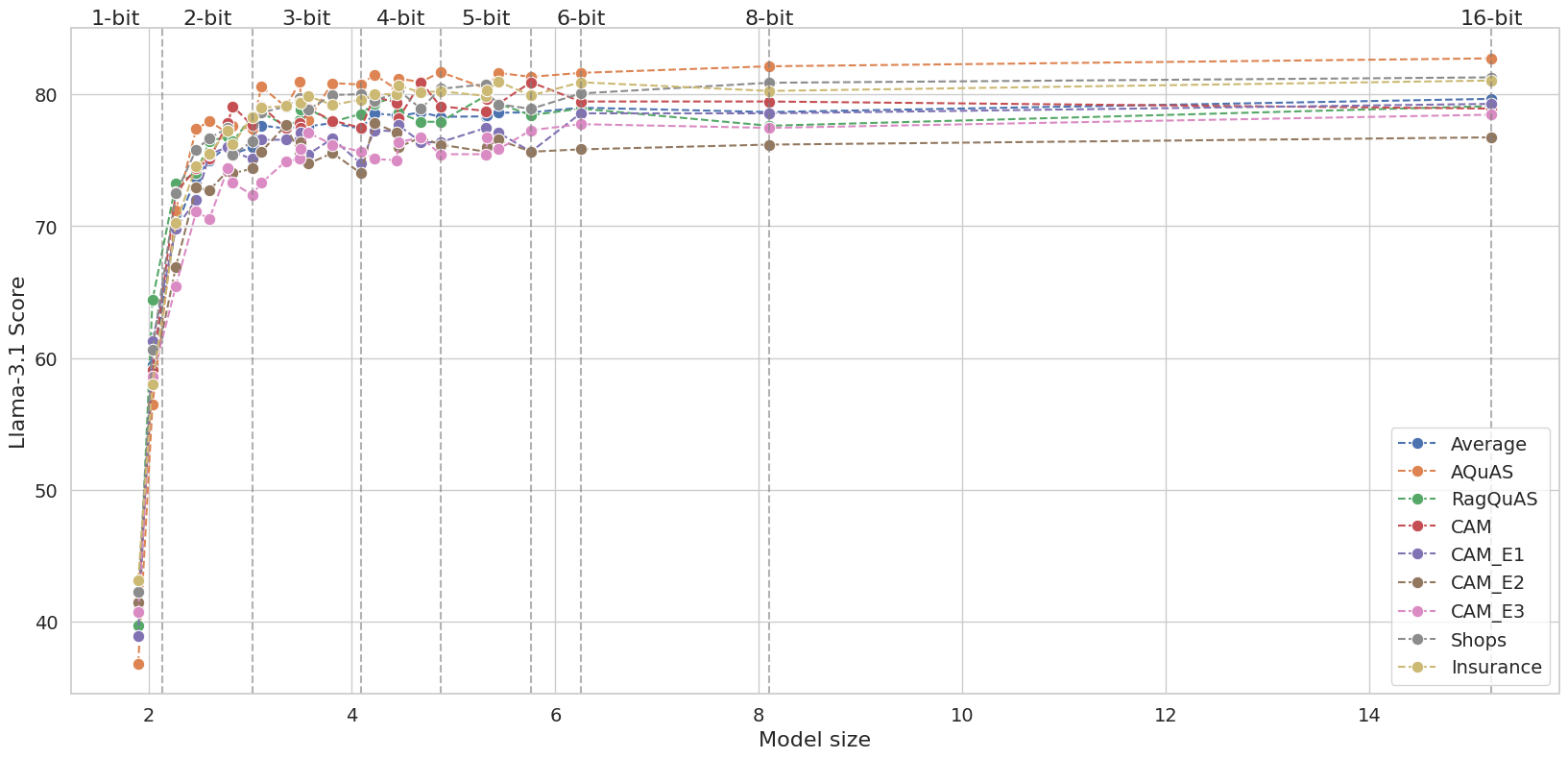}
    \caption{Line Plot of the evaluation results of RigoChat-7b-v2 across different precisions.}
    \label{fig:quantresults}
\end{figure}

The evaluations of RigoChat-7b-v2 across different quantization precisions can be found in Table \ref{tab:quantresults}. We can see that the LLM-performance increases as bit accuracy increases back to that of the original model (Figure \ref{fig:quantresults}) .Observing the performance decline as model size and precision are reduced provides valuable insight. Notably, models with as little as 3-bit precision emerge as viable and cost-effective alternatives for the specific use cases they are designed for. Additionally, models with up to 5-bit precision can be efficiently run on a standard CPU with 16 GB of RAM, making them highly accessible for various applications.

\FloatBarrier

\begin{table}[h!]
    \centering
    \renewcommand{\arraystretch}{1.35}
    \resizebox{\textwidth}{!}{%
        \begin{tabular}{rlr|c|c|c|c|c|c|c|c|c}
            \textbf{precision} & \textbf{quant-type} & \textbf{size} & \textbf{Average} & \textbf{AQuAS} & \textbf{RagQuAS} & \textbf{CAM} & \textbf{CAM\_E1} & \textbf{CAM\_E2} & \textbf{CAM\_E3} & \textbf{Shops} & \textbf{Insurance}\\
            \hline
            16-bit & - & 15.2 & 79.55 & 82.52 & 79.10 & 78.91 & 79.17 & 76.73 & 78.23 & 80.79 & 81.04 \\
            \hdashline
            8-bit & q8\_0 & 8.10 & 79.06 & 82.12 & 77.61 & 79.45 & 78.55 & 76.18 & 77.45 & 80.87 & 80.25 \\
            \hdashline
            6-bit & q6\_k & 6.25 & 79.13 & 81.62 & 78.91 & 79.45 & 78.55 & 75.82 & 77.73 & 80.06 & 80.91 \\
            \hdashline
            5-bit & q5\_1 & 5.76 & 78.50 & 81.32 & 78.41 & 80.91 & 75.64 & 75.64 & 77.27 & 78.89 & 79.90 \\
            & q5\_k\_m & 5.44 & 78.73 & 81.62 & 79.30 & 79.27 & 77.09 & 76.55 & 75.82 & 79.24 & 80.96 \\
            & q5\_0 & 5.33 & 78.63 & 80.50 & 79.90 & 79.64 & 75.82 & 76.00 & 76.73 & 80.12 & 80.30 \\
            & q5\_k\_s & 5.32 & 78.53 & 80.44 & 79.90 & 78.73 & 77.45 & 75.64 & 75.45 & 80.76 & 79.87 \\
            \hdashline
            4-bit & q4\_1 & 4.87 & 78.41 & 81.68 & 77.91 & 79.09 & 76.36 & 76.18 & 75.45 & 80.41 & 80.23 \\
            & q4\_k\_m & 4.68 & 78.57 & 80.93 & 77.91 & 80.91 & 76.36 & 76.73 & 76.73 & 78.89 & 80.13 \\
            & q4\_k\_s & 4.46 & 78.62 & 81.18 & 78.61 & 78.18 & 77.64 & 76.00 & 76.36 & 80.29 & 80.66 \\
            & q4\_0 & 4.44 & 78.62 & 79.63 & 81.09 & 79.82 & 77.64 & 76.18 & 74.55 & 80.23 & 79.82 \\
            & iq4\_nl & 4.44 & 78.61 & 81.12 & 78.51 & 78.91 & 76.73 & 78.00 & 75.45 & 79.94 & 80.20 \\
            & iq4\_xs & 4.22 & 78.79 & 81.49 & 79.30 & 79.82 & 77.27 & 77.82 & 75.09 & 79.53 & 80.00 \\
            \hdashline
            3-bit & q3\_k\_l & 4.09 & 77.58 & 80.74 & 78.51 & 77.45 & 74.73 & 74.00 & 75.64 & 80.00 & 79.59 \\
            & q3\_k\_m & 3.81 & 78.11 & 80.37 & 77.61 & 78.55 & 77.09 & 76.00 & 76.18 & 79.77 & 79.34 \\
            & q3\_k & 3.81 & 77.95 & 81.31 & 78.21 & 77.45 & 76.18 & 75.09 & 76.18 & 80.12 & 79.04 \\
            & iq3\_m & 3.57 & 77.73 & 78.06 & 78.91 & 78.91 & 75.45 & 74.73 & 77.09 & 78.83 & 79.85 \\
            & iq3\_s & 3.50 & 77.95 & 79.31 & 78.81 & 77.45 & 77.09 & 76.36 & 75.82 & 79.36 & 79.37 \\
            & q3\_k\_s & 3.49 & 77.88 & 80.93 & 78.01 & 77.82 & 76.55 & 75.45 & 75.09 & 79.71 & 79.44 \\
            & iq3\_xs & 3.35 & 77.68 & 79.06 & 77.61 & 77.45 & 76.55 & 77.64 & 74.91 & 79.07 & 79.11 \\
            & iq3\_xxs & 3.11 & 77.66 & 80.56 & 78.61 & 79.09 & 76.55 & 75.64 & 73.27 & 78.60 & 78.98 \\
            \hdashline
            2-bit & q2\_k & 3.02 & 76.23 & 77.38 & 78.31 & 77.64 & 75.09 & 74.36 & 72.36 & 76.44 & 78.27 \\
            & q2\_k\_s & 2.83 & 75.95 & 77.57 & 76.42 & 79.09 & 75.64 & 74.00 & 73.27 & 75.39 & 76.24 \\
            & iq2\_m & 2.78 & 76.37 & 77.06 & 76.82 & 77.82 & 76.00 & 74.18 & 74.36 & 77.43 & 77.26 \\
            & iq2\_s & 2.60 & 75.04 & 77.94 & 76.42 & 75.09 & 75.45 & 72.73 & 70.55 & 76.68 & 75.48 \\
            & iq2\_xs & 2.47 & 74.01 & 77.38 & 74.01 & 74.36 & 72.00 & 72.91 & 71.09 & 75.80 & 74.52 \\
            & tq2\_0 & 2.44 & 26.88 & 26.73 & 27.86 & 25.64 & 26.36 & 25.45 & 27.27 & 25.95 & 29.80 \\
            & iq2\_xxs & 2.27 & 70.24 & 71.21 & 73.23 & 72.55 & 69.82 & 66.91 & 65.45 & 72.48 & 70.23 \\
            \hdashline
            1-bit & tq1\_0 & 2.14 & 27.08 & 27.85 & 28.56 & 27.09 & 25.27 & 25.82 & 26.18 & 27.00 & 28.86 \\
            & iq1\_m & 2.04 & 59.59 & 56.45 & 64.38 & 59.09 & 61.27 & 58.36 & 58.55 & 60.64 & 58.02 \\
            & iq1\_s & 1.90 & 40.50 & 36.82 & 39.70 & 40.91 & 38.91 & 41.45 & 40.73 & 42.27 & 43.17 \\
            \hline
        \end{tabular}
    }
    \vspace{2mm}
    \caption{Evaluation results of RigoChat-7b-v2 across different precisions using llama.cpp quantizations and shorted by model size. The models size is represented in gigabytes (GB).}
    \label{tab:quantresults}
\end{table}
\FloatBarrier

\subsection{Comparison of alternative fine-tuning strategies} \label{comparisons}

Several alternative strategies to make use of the High Quality corpus from Table \ref{tab:alldatasets} and the Preferences Dataset from \ref{tab:preferences_datasets} were tested; in the following we present the most relevant ones, specifically those obtained with the Llama-3.1-8B family and the previously discussed Qwen-2.5-7B. Table \ref{tab:ablation_results} presents the average score attained by these alternative approaches in the evaluation datasets. The symbol ‘$+$’ is used to indicate several sequential steps of fine-tuning. For instance, SFT + DPO refers to the application of the Supervised Fine-Tuning algorithm followed by Direct Preference Optimization. Scores for the baseline models (no SFT or DPO applied) are also presented for reference.  It should be noted that the application of DPO directly to the base models was not tested, as it was deemed ineffective.

\begin{table}[h!]
    \centering
    \renewcommand{\arraystretch}{1.2}
        \begin{tabular}{l|c}
            \textbf{Model and algorithms} & \textbf{Llama-3.1 Score} \\
            \hline
            Qwen2.5-7B & $68.32$ \\
            Qwen2.5-7B + SFT & $69.23$ \\
            Qwen2.5-7B + SFT + DPO & $72.29$ \\
            Qwen2.5-7B-Instruct & $77.17$ \\
            Qwen2.5-7B-Instruct + SFT & $70.41$\\
            Qwen2.5-7B-Instruct + SFT + DPO & $72.43$ \\
            \textbf{Qwen2.5-7B-Instruct + DPO} & \textbf{79.55} \\
            Llama-3.1-8B & $66.16$ \\
            Llama-3.1-8B + SFT & $69.04$ \\
            Llama-3.1-8B + SFT + DPO & $71.11$ \\
            Llama-3.1-8B-Instruct & $76.55$ \\
            Llama-3.1-8B-Instruct + SFT & $70.36$ \\
            Llama-3.1-8B-Instruct + SFT + DPO & $72.13$ \\
            Llama-3.1-8B-Instruct + DPO & $78.38$ \\
            \hline
        \end{tabular}
    \vspace{2mm}
    \caption{Average Llama-3.1 Score in the set of evaluations of all the experiments.}
    \label{tab:ablation_results}
\end{table}
\FloatBarrier

The training parameters used in DPO are presented in Table \ref{tab:dpotraining}, while Table \ref{tab:sfttraining} details those for Supervised Fine Tuning. Additionally, LoRA was employed with the configuration outlined in Table \ref{tab:loraconfig}. Notably, applying our training approach to the Llama-3.1-8B-Instruct model also demonstrates improvements, further validating the approach taken to adapt LLMs to use cases efficiently and effectively.

\begin{table}[h!]
\centering
\renewcommand{\arraystretch}{1.25}
\begin{tabular}{c|c}
\textbf{DPOConfig argument} & \textbf{Value} \\
\hline
num\_train\_epochs & $2$ \\
eval\_steps & $50$ \\
save\_steps & $150$ \\
save\_total\_limit & $5$ \\
per\_device\_train\_batch\_size & $2$ \\
per\_device\_eval\_batch\_size & $2$ \\
gradient\_accumulation\_steps & $16$ \\
learning\_rate & $2\cdot 10^{-4}$ \\
max\_length & $8192$ \\
gradient\_checkpointing & True \\
weight\_decay & $0.001$ \\
optim & adamw\_torch \\
lr\_scheduler\_type & cosine \\
\hline
\end{tabular}
\vspace{2mm}
\caption{Configuration for the SFT training. Parameters not shown here used  the default values of the library.}
\label{tab:sfttraining}
\end{table}
\FloatBarrier

\subsection{General performance of RigoChat-7b-v2} \label{general_performance}

After demonstrating that RigoChat-7b-v2 achieves improvements in certain tasks related with question answering with contexts in Spanish, it is essential to assess whether for a broader selection of tasks it remains as competent as Qwen2.5-7B-Instruct (the original base model), or if some kind of catastrophic forgetting or heavy task-focusing has taken place. To this end, Table \ref{tab:rigochatvsqwen} presents a comparison of results between RigoChat-7b-v2 and Qwen2.5-7B-Instruct across some widely used datasets for LLM evaluation. Although some noticeable drops in performance can be observed for the DROP, HumanEval and IfEval datasets,  the rest of tasks present similar performance for both models, with a significant improvement taking place for the Spanish MMMLU dataset. This confirms that the proposed procedure has succeeded in aligning the base model to the Spanish language, while retaining most of the capabilities of the base model for the English language and general-purpose tasks.

\begin{table}[h!]
\centering
\renewcommand{\arraystretch}{1.4}
\begin{tabular}{lcc}
    \toprule
    \textbf{Datasets} & \textbf{RigoChat-7b-v2} & \textbf{Qwen2.5-7B-Instruct} \\
    \midrule
    \multicolumn{3}{c}{\centering \textbf{General}} \\
    \hdashline
    MMLU \cite{DBLP:journals/corr/abs-2009-03300} & $74.07$ & $74.24$ \\
    MMMLU \cite{DBLP:journals/corr/abs-2009-03300} (ES) & $62.95$ & $60.44$ \\
    MMLU-Pro \cite{wang2024mmluprorobustchallengingmultitask} & $55.70$ & $56.20$  \\
    \midrule
    \multicolumn{3}{c}{\textbf{Reasoning}} \\
    \hdashline
    GPQA \cite{rein2023gpqagraduatelevelgoogleproofqa} & $33.84$ & $32.83$ \\
    HellaSwag \cite{zellers2019hellaswag} & $84.78$ & $85.15$ \\
    DROP \cite{dua2019dropreadingcomprehensionbenchmark} & $62.14$ & $66.20$ \\
    \midrule
    \multicolumn{3}{c}{\textbf{Math}} \\
    \hdashline
    MATH \cite{hendrycks2021measuringmathematicalproblemsolving} & $68.42$ & $69.16$ \\
    GSM8K \cite{cobbe2021trainingverifierssolvemath} & $79.83$ & $80.52$ \\
    \midrule
    \multicolumn{3}{c}{\textbf{Coding}} \\
    \hdashline
    HumanEval \cite{chen2021evaluating} & $78.66$ & $81.71$ \\
    LiveCodeBench \cite{jain2024livecodebenchholisticcontaminationfree} & $15.38$ & $12.44$ \\
    \midrule
    \multicolumn{3}{c}{\textbf{Aligment}} \\
    \hdashline
    IfEval \cite{zhou2023instructionfollowingevaluationlargelanguage} & $75.42$ & $79.6$  \\
    \bottomrule
\end{tabular}
\vspace{2mm}
\caption{General performance of RigoChat-7b-v2 and Qwen-2.5-7B-Instruct across classical LLM benchmarks.}
\label{tab:rigochatvsqwen}
\end{table}
\FloatBarrier

These evaluations have been carried out with the open-source evaluation tool OpenCompass \cite{2023opencompass}.

\section{Conclusions and Future Work}

Leveraging limited hardware resources, our study demonstrates that robust performance is achievable—even surpassing state-of-the-art models in certain scenarios. Crucially, our results emphasize that high-quality and representative data are as important as abundant computational power. By meticulously curating our datasets, we effectively align a multilingual, open-source model to perform more effectively in Spanish, thereby democratizing access to advanced AI while enhancing privacy, reducing dependency on third-party providers, and optimizing resource efficiency.

The proposed approach to augment data by obtaining responses from different models on threads of interest has been shown to be an effective distillation method if these responses are ranked and DPO is applied. Furthermore, the model obtained using this approach, RigoChat-v2, does not lose general performance in those tasks for which its base model was initially designed, as shown in the evaluations of classical tasks in \ref{general_performance}.

For future research focused on improving LLMs, we plan to:

\begin{itemize}
    \item \textbf{Expand and enhance data quality:} Make a substantial investment in acquiring and curating high-quality datasets, as they serve as long-lasting assets that can be reused in the future. Unlike models and GPUs, which can become outdated quickly and require repeated investments, high-quality data maximize the potential of available hardware while offering a more sustainable and cost-effective strategy.
    
    \item \textbf{Develop robust evaluation methods:} Advance methodologies to automate human evaluation simulation, Data Augmentation and further maximize data utilization.

    \item Integrate advanced training techniques to \textbf{take advantage of all data} presented in \ref{tab:alldatasets}. Specifically, apply Group Relative Policy Optimization (GRPO) \cite{shao2024deepseekmathpushinglimitsmathematical} on tasks where outcome quality can be easily measured, in conjunction with our best-performing DPO-trained model on our large-scale instruction corpus. This approach could effectively transfer the high-quality responses from the preference corpus to the extensive set of threads available.
\end{itemize}

These steps aim to reinforce the pivotal role of data quality in achieving superior model performance and to drive continued innovation in resource-efficient AI.

Finally, the significant progress in high-quality massive multilingual corpora, such as FineWeb-2 \cite{penedo2024fineweb-2}, combined with the stagnation in the publication of Natural Language Understanding (NLU) models until the release of ModernBERT \cite{warner2024smarterbetterfasterlonger}—a state-of-the-art model incorporating the latest innovations—makes it feasible to train a Spanish-specific NLU model. This approach follows the path we previously took with RigoBERTa \cite{serrano2022rigobertastateoftheartlanguagemodel}.

\bibliographystyle{unsrt} 
\bibliography{references}  

\end{document}